\documentclass[pmlr]{jmlr}% new name PMLR (Proceedings of Machine Learning Research)

\usepackage{amsmath,amsfonts,bm}

\def\1{\bm{1}}

\def\ru{{\textnormal{u}}}
\def\rv{{\textnormal{v}}}
\def\rw{{\textnormal{w}}}

\def\mA{{\bm{A}}}

\def\mH{{\bm{H}}}

\DeclareMathAlphabet{\mathsfit}{\encodingdefault}{\sfdefault}{m}{sl}
\SetMathAlphabet{\mathsfit}{bold}{\encodingdefault}{\sfdefault}{bx}{n}

\def\gA{{\mathcal{A}}}

\def\gL{{\mathcal{L}}}

\def\sN{{\mathbb{N}}}

\newcommand{\E}{\mathbb{E}}

\newcommand{\R}{\mathbb{R}}

\usepackage{mathdots}
\usepackage{pmat}

\usepackage{longtable}% for long tables

\usepackage{booktabs}
\usepackage[load-configurations=version-1]{siunitx} % newer version
 \usepackage{booktabs}       % professional-quality tables
\usepackage{amsfonts}       % blackboard math symbols
\usepackage{nicefrac}       % compact symbols for 1/2, etc.
\usepackage{microtype}      % microtypography
\usepackage{hyperref}

\usepackage{epstopdf} 

\usepackage{algorithm}
\usepackage[noend]{algorithmic}

\usepackage{float}

\usepackage{lipsum}
\usepackage{url}
\usepackage{stmaryrd}

\usepackage[utf8]{inputenc}
\usepackage{tikz,siunitx}
\usetikzlibrary{positioning}
\usetikzlibrary{calc}

\usetikzlibrary{decorations.pathreplacing}
\usetikzlibrary{cd,arrows,shapes,shapes.misc,intersections}

\usepackage{cancel}
\usepackage{siunitx}
\usepackage{array}
\usepackage{multirow}
\usepackage{amssymb}
\usepackage{gensymb}
\usepackage{mathdots}
\usepackage{graphicx}
\usepackage{bm}
\usepackage{esvect}
\usepackage[algo2e]{algorithm2e}
\allowdisplaybreaks
\usepackage[symbol]{footmisc}
\usepackage{float}

\theorembodyfont{\upshape}
\theoremheaderfont{\scshape}
\theorempostheader{:}
\theoremsep{\newline}

\newtheorem*{proof}{Proof}

\jmlrvolume{1}
\jmlryear{2022}
\jmlrworkshop{Learning and Automata Workshop}

\usepackage{todonotes}

\author{
\Name{Kaiwen Hou} \Email{kaiwen.hou@columbia.edu}\\
  \addr Tsinghua University \& Mila
\AND
\Name{Guillaume Rabusseau} \Email{grabus@iro.umontreal.ca}\\
    \addr DIRO, Universit\'e de Montr\'eal \& Mila, CIFAR AI Chair}

\title{Spectral Regularization: \\
an Inductive Bias for Sequence Modeling}

\begin{document}

\maketitle

\begin{abstract}%   <- trailing '%' for backward compatibility of .sty file
% Machine learning tasks extensively involve regularization to prevent overfitting, various forms of which strive for different notions of simplicity. This paper presents spectral regularization techniques, which attaches a unique inductive bias to sequence modeling tasks based on an intuitive concept of simplicity defined in Chomsky hierarchy. From prior works revealing fundamental associations between second-order RNNs and weighted automata and between Hankel matrices and regular grammars, we utilize the trace norm of the Hankel matrix, being a sub-differentiable convex relaxation of its rank, as the spectral regularizer. Considering that the Hankel matrix is infinite, we propose an unbiased Russian Roulette estimator to estimate its trace norm. Ultimately, we demonstrate experimental results which exhibit the prowess of spectral regularization.
Various forms of regularization in learning tasks strive for different notions of simplicity. This paper presents  a spectral regularization technique, which attaches a unique inductive bias to sequence modeling based on an intuitive concept of simplicity defined in the Chomsky hierarchy. From fundamental connections between Hankel matrices and regular grammars, we propose to use the trace norm of the Hankel matrix, the tightest convex relaxation of its rank, as the spectral regularizer. To cope with the fact that the Hankel matrix is bi-infinite, we propose an unbiased stochastic estimator for its trace norm. Ultimately, we demonstrate experimental results on Tomita grammars, which exhibit the potential benefits of spectral regularization and validate the proposed stochastic estimator.

\end{abstract}


\section{Introduction}
% \gr{TODO...}
% Chomsky hierarchy categorizes functions over sequences into four different levels of complexity, the simplest of which is regular grammar \citep{1056813}. To encourage a sequential model like RNN (Recurrent Neural Network) to learn simple functions, i.e., functions that appear lower in Chomsky hierarcky, this paper introduces a novel inductive bias through spectral regularization. 
% % \textcolor{red}{why learn functions low in hierarchy? any RNN or just put the right bias in regular grammars?} 
% Based on recent work showing fundamental connections between second-order RNNs and weighted automata \citep{Rabusseau2019ConnectingWA}, and the close relationship between Hankel matrices and regular grammars, this paper uses the trace norm of Hankel matrix as a sub-differentiable convex relaxation of the rank for regularizing the loss used for training various sequential model architectures.

The fundamental principle of \emph{regularization} is at the heart of many machine learning algorithms and models. Informally speaking, regularization refers to the idea of adding a penalty term to the loss function optimized by a learning model in order to encourage learning \emph{simple} functions. In particular, in regularized empirical risk minimization, the objective is to find the hypothesis $h$ minimizing $\gL(h,D)+\lambda \Omega(h)$ where $\gL(h,D)$ denotes the empirical risk on a dataset $D$, $\Omega(h)$ is a penalty term that penalizes complex functions and $\lambda$ is an hyper-parameter controlling the tradeoff between fitting the data and $h$ being a "simple" hypothesis. Examples of regularization functions $\Omega$ include the $\ell_2$ or $\ell_1$ norm of the weight vector of a linear model, the degree of a polynomial model, the rank or the trace norm of the user-item matrix in a collaborative filtering task, etc. 

In this work, we propose a novel regularization technique for sequential models. While there are many natural notions of simplicity for functions defined over vector spaces~(e.g., sparsity, smoothness, etc.), defining a notion of simplicity suited for functions defined over sequences can be more tedious due to the discrete and sequential nature of the data arising in tasks such as language modelling. One such notion of simplicity naturally arises from the so-called Chomsky hierarchy, which categorizes functions over sequences into four different levels of complexity, the simplest of which is regular grammars~\citep{1056813}. To encourage a sequential model such as an RNN (Recurrent Neural Network) to learn simple functions, i.e., functions that appear lower in Chomsky hierarchy, we introduce a novel inductive bias through \emph{spectral regularization}. 

In order to encourage learning such simple functions, we leverage a fundamental result relating the rank of the Hankel matrix of a function $f$ to the minimal number of states of a weighted finite automaton computing $f$~\citep{fliess1974matrices,carlyle1971realizations}. Functions computed by weighted finite automata corresponds to regular weighted languages, i.e., functions that are low in the Chomsky hierarchy. The idea behind spectral regularization is to encourage learning models whose Hankel matrix are approximately low rank. To do so, the spectral regularization is defined as the trace norm of the Hankel matrix. Using the trace norm instead of the rank of the Hankel matrix offers two advantages: (i) the trace norm is the tightest convex relaxation of the rank and is differentiable, allowing one to use automatic differentiation techniques to use the spectral regularization when training black box neural network sequence models, and (ii) the trace norm can be seen as a "soft" version of the rank, allowing learned models to only be \emph{approximately} low rank, whereas a hard rank constraint would be too strong and forces the learned functions to be regular. The spectral regularization can thus incorporate a natural inductive bias towards regular functions in the training of any black box differentiable model. 

A key technical challenge in implementing the proposed spectral regularization resides in the fact the Hankel matrix is a bi-infinite matrix whose trace norm cannot be explicitly computed. To address this issue we propose a Russian Roulette estimator to design a stochastic unbiased estimator of the Hankel matrix, whose trace norm is lower bounded~(in expectation) by the trace norm of the Hankel matrix itself. We thus plug in the realizations of the Russian Roulette estimator in the minimization objective at each mini-batch  in place of the actual trace norm of the Hankel matrix. 

We provide a simple experimental study on Tomita grammars~\citep{tomita1982dynamic} to illustrate the potential benefits of the spectral regularization.

\section{Preliminaries}
% {\color{blue}
% \begin{definition}[\textbf{Monoid}]
% A monoid is a triple $(\Sigma,\boldsymbol\cdot,\Lambda)$, where $\Sigma$ is a set with an associative binary operation $\boldsymbol\cdot: \Sigma\times\Sigma\to\Sigma$ (so that $(\Sigma,\boldsymbol\cdot)$ is a semigroup) and an identity element $\Lambda$.
% \end{definition}

% Let $\Sigma$ denote the alphabet set of a language, and $\boldsymbol\cdot$ the operation of string concatenation which maps some arbitrary $\ru,\rv\in\Sigma$ to $\ru\rv:=\ru\boldsymbol\cdot\rv\in\Sigma$. 
% Obviously $\boldsymbol\cdot$ is associative, since
% $\forall \ru,\rv,\rw\in\Sigma: (\ru\rv)\rw=\ru(\rv\rw)$. Together with the empty string being the identity element $\Lambda$, $(\Sigma,\boldsymbol\cdot,\Lambda)$ forms a monoid.
% We denote by $\Sigma^*$ the Kleene star of $\Sigma$, i.e. \textcolor{red}{sigma star is the set of monoid instead of sigma}

% also called free monoid.

% $\varepsilon$
% $\epsilon$ %empty

% Here  $\Sigma^*:=\bigcup\limits_{n=0}^{\infty}\Sigma^n$ represents the set of all sequences in $\gG$.}

Let $\Sigma$ be a finite nonempty set, also known as an alphabet. We denote by $\Sigma^*$ the free monoid over $\Sigma$, where string concatenation is the binary operation and the empty string in the singleton set $\Sigma^0:=\{\epsilon\}$ serves as the unique unit element. Intuitively, $\Sigma^*$ refers to the set of all finite sequences (or words) generated by $\Sigma$:
            \begin{equation*}
            \Sigma^*:=\bigcup\limits_{n=0}^{\infty}\Sigma^n.
            \end{equation*} 
For two sequences $\ru,\rv\in\Sigma^*$, we use $\ru\rv$ to denote the concatenation of $\ru$ and $\rv$. The length of a sequence $\rw\in\Sigma^*$ is denoted as $|\rw|$. Finally, a grammar, or language, over $\Sigma$ is a subset of $\Sigma^*$.

One of the simplest class of languages is the set of \emph{regular languages}, which are languages that can be computed by deterministic finite automata. Regular languages forms  the simplest class of languages in the so-called Chomsky hierarchy~\cite{}. In this work, we are interested in real-valued functions over $\Sigma^*$, sometimes called weighted languages. Such functions are of crucial interest for machine learning applications on sequence data such as language modeling. The Chomsky hierarchy easily extends to weighted languages using the weighted counterparts of the finite state machines used in the classical hierarchy. In particular, the simplest class of such functions is the set of regular functions~(sometimes called rational, or recognizable), which are functions that can be computed by weighted automata. 

\begin{definition}[\textbf{Weighted Finite Automaton}]
A weighted finite automaton (WFA) with $n$ states is a tuple $\gA=(\alpha,\{\mA^\sigma\}_{\sigma\in\Sigma},\omega)$, where $\alpha\in\R^n$ is the initial weight vector, $\omega\in\R^n$ the final weight vector, and $\mA^\sigma\in\R^{n\times n}$ is the transition matrix for each symbol $\sigma\in\Sigma$.
A WFA computes a function $f_\gA: \Sigma^*\rightarrow\R$ that maps any sequence $\ru=\ru_1\ru_2\cdots\ru_k\in\Sigma^*$ to $f_\gA(\ru)
= \alpha^T\mA^\ru\omega$, where $\mA^\ru:=\mA^{\ru_1}\mA^{\ru_2}\cdots\mA^{\ru_k}$. 
\end{definition}
% WFA offers a rigorous way to categorize subsets of $\Sigma^*$ as Chomsky hierarchy presents. The following theorem characterizes regular grammar, which appears lowest in the hierarchy.
% \begin{theorem}
% A grammar is regular if and only if it can be recognized by a WFA.
% \end{theorem}

% Alternatively, the properties of Hankel matrix suggests another way to categorize grammars.
It is worth briefly mentioning that any regular language is the support of a rational function~(however, surprisingly, the converse is not true, see, e.g., Chapter 4, Section 6 in~\cite{droste2009handbook}). 

In this work, we will design a regularization scheme for sequential models that will favour functions that are close to the class of rational functions~(i.e., low on Chomsky's hierarchy). In order to do so, we need a quantitative measure of the "rationality" of a function. We will see that the spectrum of the so-called Hankel matrix is a good candidate for this purpose.

\begin{definition}[\textbf{Hankel Matrix}]
For a given function $f:\Sigma^*\rightarrow\R$, its Hankel matrix $\mH_f\in\R^{\Sigma^*\times\Sigma^*}$ is the infinite matrix with entries $(\mH_f)_{\ru,\rv}=f(\ru\rv)$ for $\ru,\rv\in\Sigma^*$.
\end{definition}

The following classical theorem shows the fundamental~(striking) relation between the Hankel matrix of a function and its "rationality". 
\begin{theorem}[{\cite{fliess1974matrices,carlyle1971realizations}}] \label{thm:fliess}
For any function $f:\Sigma^*\rightarrow\R$, $rank(\mH_f)$ is equal to the minimal number of states of a WFA computing $f$. In particular, a function $f$ is regular if and only if its Hankel matrix has finite rank. 
\end{theorem}
% \begin{corollary}
% A grammar is regular if and only if there exists a function  $f:\Sigma^*\rightarrow\R$ such that $rank(\mH_f)<\infty$.
% \end{corollary}

We will see in the next section how this result can be leveraged to design a regularization technique to favour simpler model during learning. 

\section{Spectral Regularization}
In this section we propose the trace norm of the Hankel matrix as a natural spectral regularization for black box sequential models and show how to efficiently compute stochastic approximation of this regularization term for training through back-propagation. 

% So far, we notice that learning a regular grammar corresponds to constructing a low-rank Hankel matrix, and that the trace norm is a convex envelope of the rank function ${\displaystyle {rank}(\mH)}$, which motivates the following regularization technique.

% The Hankel matrix %$\mH(\gG)\in\R^{\sP\times\sS}$ 
% $\mH(\gG)\in\R^{\Sigma^*\times\Sigma^*}$ w.r.t a regular grammar $\gG$ of alphabet $\Sigma$, prefixes $\sP\subset\Sigma^*$, and suffixes $\sS\subset\Sigma^*$ is given by
% \begin{equation*}
%     \mH_{\ru,\rv} = P(\overline{\ru\rv}) \quad \forall \ru\in\sP,\rv\in\sS,
% \end{equation*}
% denoting the probability a sequence $\overline{\ru\rv}$ appears in $\gG$.
%\textcolor{red}{regular grammar => H or the other way?}

% Suppose an RNN is trained to learn the underlying pattern of a sequence, which is generated according to a regular grammar. The following defines the loss under spectral regularization based on Hankel matrix $\mH$.
\subsection{Motivation and Definition}
A naive idea to leverage Theorem~\ref{thm:fliess} for regularization would be to enforce the Hankel matrix of the learned model to be low rank. However, this approach has two drawbacks. First, optimization under low rank constraints is known to be computationally hard. Second, such a constraint would be too strong: we want to incorporate an inductive bias towards simple functions in the learning process, but we do not want to actually enforce the learned function to be regular. In some sense, we want a softer version of the rank of the Hankel matrix which would also consider functions that can be well approximated by regular functions~(i.e., functions whose Hankel matrix is approximately low rank) as simple. 

Enforcing the trace norm~(or nuclear norm) of the Hankel matrix to be small, instead of directly enforcing the rank to be small, will solve (to some extent) both of these issues. Indeed, the trace norm~(which is the sum of the singular values) is the tightest convex relaxation of the matrix rank~\citep{fazel2001rank} which naturally represents a soft version of the notion of rank. The trace norm of the Hankel matrix has actually been previously leveraged for this purpose in the context of learning~\citep{balle2012local}. Using the trace norm of the Hankel matrix as a way to regularize models for sequence tagging was also previously explored in~\citep{quattoni2014spectral}.

We formally introduce this regularization technique in the following definition. 

\begin{definition}[\textbf{Spectral Regularization}]
% An RNN without regularization has a loss $\Tilde{\gL}$, then the spectral regularization corresponds to the following minimization problem
Let $f_\theta:\Sigma^* \to \mathbb{R}$ be the function computed by a model with parameters $\theta$ and let $\tilde{\gL}(\theta)$ be the loss function associated with this model. Spectral regularization corresponds to the following minimization problem:
\begin{equation}\label{specReg}
 \min_{\theta} \Tilde{\gL}(\theta) + \lambda ||\mH_{f_\theta}||_*,
\end{equation}
where $\lambda$ is the regularization coefficient, and the trace norm $||\mH_{f_\theta}||_*$ is the spectral regularizer (or spectral loss).
\end{definition}

Note that the previous definition does not make any assumptions on the class of models considered. One particular class of interest is the one of functions computed by recurrent neural networks, for which we would ideally want to, at the same time, benefit from their remarkable expressiveness while still steering the learning process towards functions that are, in some sense, low on the Chomsky hierarchy. 
In particular, when an RNN is used for sequential probabilistic modeling~(i.e. trained to predict the probabilities of next symbol given a sequence), $f_\theta$ would denote the underlying probability distribution over $\Sigma^*$, i.e., $f_{\theta}(\ru_1\ru_2\cdots\ru_k) = P(\ru_1\ru_2\cdots\ru_k) = P(\ru_1)P(\ru_2\mid \ru_1) \cdots P(\ru_k\mid \ru_1\cdots\ru_{k-1})$.

\subsection{Russian Roulette Estimator}
It is clear that the optimization problem in Eq.~\eqref{specReg} can not be solved easily. To start with, the Hankel matrix is infinite! In order to tackle this optimization problem, we will make use of the so-called Russian Roulette estimator~\cite{} which allows one to stochastically approximate an infinite series with random realization of partial sums. 

\begin{definition}[\textbf{Russian Roulette Estimator}; \cite{Kahn}] \label{def:RR}
Given a convergent series $S=\sum\limits_{k=0}^{\infty}\alpha_k$, a Russian Roulette estimator of $S$ is given by $\hat{S}:=\sum\limits_{i=0}^{\tau}\frac{\alpha_i}{P(\tau\geq i)}$, where $\tau\geq 0$ is a random variable with support over all nonnegative integers.
%satisfying $P(\tau\geq n)> 0$ for all $n$ and $\sum\limits_{k=1}^{\infty}P(\tau\geq k)<\infty$.
\end{definition}

% \gr{we need a statement with ref for the fact that this is an unbiased estiamtor (under what conditions???)}
Note that in this definition we do not require $\alpha_k$'s to be scalars. Instead, they could stand for vectors, matrices, tensors, or some abstract objects with well-defined component-wise addition.

\begin{theorem}[{\cite{NEURIPS2019_5d0d5594}; Lemma 3; \cite{lyne2015russian}}] \label{thm:chen}
If $P(\tau\geq n)>0 \ \forall n>0$ and the series $S$ is absolutely convergent, then $\hat{S}$ given in Definition~\ref{def:RR} is an unbiased estimator of $S$, i.e.,
    $\E \hat{S} = S$.
\end{theorem}

Although the Russian Roulette Estimator is unbiased under mild assumptions, its variance might be large or even unbounded with an ill-chosen random variable $\tau$ \citep{DBLP:journals/mcma/McLeish11,DBLP:conf/icml/BeatsonA19}.

\subsection{Stochastic Estimator for the Trace Norm of the Hankel Matrix}
In order to leverage the Russian Roulette estimator for the trace norm of the Hankel matrix, we need to express the Hankel matrix as an infinite sum. We propose one way convenient way to do this in the following theorem. 

\begin{theorem}
Let $f:\Sigma^*\to\R$. For any $i\in\mathbb{N}$, let $\mH^{(i)}_f \in \mathbb{R}^{\Sigma^* \times \Sigma^*}$ be defined by 
$$(\mH^{(i)}_f)_{\ru,\rv} = \begin{cases} 
f(\ru\rv) & \text{ if } |\ru\rv| = i\\
0 & \text{otherwise}\end{cases}
$$
for all $\ru,\rv\in\Sigma^*$. Then $\mH_f = \sum\limits_{i=0}^\infty \mH^{(i)}_f$.
\end{theorem}

Although the $\mH^{(i)}_f$'s defined above are infinite matrices, each of them only contains a finite number of nonzero elements. We can thus construct the Russian Roulette estimator of $\mH_f$ as
% \begin{equation}
%     \mH_\tau=\sum\limits_{k=0}^{\tau}\frac{\mH_f^{(k)}}{P(\tau\geq k)}.
% \end{equation}
\begin{equation}\label{specReg_tau}
    \mH_\tau=\sum\limits_{i=0}^{\tau}\frac{\mH^{(i)}_f}{P(\tau\geq i+1)}
\end{equation}
where $\tau$ is a random variable taking its values in $\mathbb{N}$ such that $P(\tau\geq n)> 0$ for all $n$.% and $\sum\limits_{k=1}^{\infty}P(\tau\geq k)<\infty$.

As mentioned previously, even though $\mH_\tau$ still is an infinite matrix, it only has a finite number on non-zeros entries  for any integer $\tau$. Thus, informally, the trace norm of the infinite matrix $\mH_\tau$ is equal to the trace norm of its smallest sub-block containing no columns or rows entirely filled with $0$'s, which is a finite sub-block whose trace norm can be computed in polynomial time. We now formalize this intuition. We start by showing that the Russian Roulette estimator of the Hankel matrix is unbiased.

\begin{theorem}\label{thm:RR.hankel.unbiased}
Let $f:\Sigma^*\to\R$. The estimator $\mH_\tau$ defined in Eq.~\eqref{specReg_tau} is an unbiased estimator of $\mH_f$, i.e., $\E_\tau [\mH_\tau] = \mH_f$.
\end{theorem}
\begin{proof}
%\gr{TODO}
For some $\ru, \rv \in \Sigma^*$, we notice from Eq.~\eqref{specReg_tau} that $$(\mH_\tau)_{\ru, \rv} = \begin{cases} 
\frac{(\mH^{(i)}_f)_{\ru,\rv}}{P(\tau\geq i+1)}  & \text{ if } \tau\geq k+1\\
0 & \text{otherwise}\end{cases}
$$ for some $k\in\sN$, since the RHS of Eq.~\eqref{specReg_tau} contributes at most one term for an entry in LHS. Then
$$
\E [(\mH_\tau)_{\ru, \rv}] = \E [ \frac{(\mH^{(i)}_f)_{\ru,\rv}}{P(\tau\geq i+1)}\1_\mathrm{\tau\geq k+1}] = \frac{(\mH^{(i)}_f)_{\ru,\rv}}{P(\tau\geq k+1)}P(\tau\geq k+1) = (\mH^{(i)}_f)_{\ru,\rv}.
$$
Therefore, each entry of $\mH_\tau$ is unbiased to estimate the corresponding entry in $\mH_f$.

$\hfill\blacksquare$
\end{proof}

We showed that the infinite Hankel matrix of a function can be expressed as an infinite sum of matrices with a finite number of non-zero entries, allowing us to construct a Russian Roulette estimator of the Hankel matrix which can be computed efficiently. But we are interested in the trace norm of the Hankel matrix in the objective we wish to minimize in Eq.~\eqref{specReg}. It remains to show that the trace norm of the Russian Roulette estimator of the Hankel matrix is a good stochastic estimator of the trace norm of the Hankel matrix itself.

% According to Theorem~\ref{boundE} in the following, we consider a more relaxed minimization problem with spectral regularization than in (\ref{specReg}):
% \begin{equation}\label{specRegWithExp}
%      \min_{\theta} \Tilde{\gL}(\theta) + \lambda \E||\mH_\tau||_*.
% \end{equation}

\begin{theorem}\label{boundE}
% Let $\tau$ be a random variable following a geometric distribution.
For any $\theta$, we have that 
\begin{equation}
   \Tilde{\gL}(\theta) + \lambda ||\mH_{f_\theta}||_* \leq \Tilde{\gL}(\theta) + \lambda \E [||\mH_\tau||_*],
\end{equation}
where $\mH_\tau$ is the Russian Roulette estimator defined in Eq.~\eqref{specReg_tau}.
\end{theorem}
\begin{proof}
It suffices to show that $||\cdot||_*$ is a convex operator, and then the claim follows by Jensen's inequality. The convexity of $||\cdot||_*$ follows naturally from the triangle inequality of the trace norm $||\mu\mH_1+(1-\mu)\mH_2||_*\leq\mu||\mH_1||_*+(1-\mu)||\mH_2||_*$ for any $\mu\in[0,1]$. Combining Jensen's inequality with Theorem~\ref{thm:RR.hankel.unbiased} we have that $ \E [||\mH_\tau||_*] \geq  ||\E [\mH_\tau]||_* = ||\mH_{f_\theta}||_* $.
% E is unital operator

$\hfill\blacksquare$
\end{proof}

The previous theorem shows that we can efficiently compute a stochastic approximation of the trace norm of the Hankel matrix, through which we can use the back-propagation algorithm to train any differentiable black-box model. In the next section, we implement this regularization technique to train RNNs on a synthetic language modeling task.

\section{Experiments}
% \gr{TODO...
% \begin{itemize}
%     \item Introduce Tomita grammars 3-6
%     \item describe model architecture (number of hidden neurons, embedding layer...)
%     \item dataset division 
%     \item Introduce the implementation details: optimization algorithm (Adam, cite!), early stopping on a validation set, simple scheduler to reduce learning rate on plateau, batch size = 32, ...)
%     \item We compare with and without regularization. When using regularization lambda is chose according to a validation set. 
%     \item we compare with a naive biased estimator... 
%     \item metric for training and test
%     \item conclusion: spectral regularization marginally helps for tom 4-5-6 on small dataset sizes. No clear winner between biased and RR. Maybe results would be better on other tasks (classification) oir other datasets, to be explored in future work.
% \end{itemize}

% }

We conduct experiments to validate that spectral regularization imposes an inductive bias for sequence modeling. In particular, we focus on synthetic data generated according to Tomita grammars \#3 to \#6 defined in Table~\ref{tab:xp_tomita}, which is a benchmark study for grammatical inference~\citep{tomita1982dynamic,bengio1994approach}. As shown in the table, all these grammars are some subsets of $\Sigma^*$, where the binary alphabet is $\Sigma:=\{0,1\}$.

\begin{table}[H]
\centering
\begin{tabular}{|c|l|}
    \hline
    Tomita Grammars &  Definitions \\
    \hline\hline
    \#3 & not containing $1^{2n+1}0^{2m+1}$ as a substring \\
    \hline
    \#4 & not containing $000$ as a substring \\
    \hline
    \#5 & containing even number of $01$'s and $10$'s \\
    \hline
    \#6 & (number of 0's $-$ number of $1$'s) is a multiple of 3 \\
    \hline
\end{tabular}
\caption{Definitions of Tomita grammars \#3 to \#6.}
\label{tab:xp_tomita}
\end{table}
The training dataset for each grammar include synthetic sequences up to length 12 in that grammar. 20\% of the training set is split out as the validation set. We also use a test dataset consisting of sequences of exact length 12 and disjoint with the training dataset. % respectively, to verify how well the model is fitting the data and how well it can generalize to unforeseen data that still belong to the grammar.

We consider an RNN with one embedding layer of $|\Sigma|+2$ neurons~(we use two additional symbols to mark the start and end of sequences) and one hidden layer of 50 neurons, which has been just expressive enough for our training data. NLL (Negative Log-Likelihood) loss is used for both training and reporting performances of grammatical inference on the three test sets. The loss minimization is based on the \emph{Adam} optimizer~\citep{diederik2014adam} with an initial learning rate of 0.01 and a batch size of 32. Moreover, early stopping and a simple scheduler to reduce learning rate on detected plateaus of validation loss are adopted. 

In our experiments, we compare the test NLL when training without spectral regularization versus that with spectral regularization for different size of training data sampled from the given training set. The latter chooses the hyperparameter $\lambda$ according to validation NLL. In each mini-batch, we randomly draw $\tau\sim \text{Geometric}(0.2)$ (stopping probability is 0.2) to construct the Russian Roulette estimator of the Hankel matrix. To check the significance of the unbiased Russian Roulette estimator in spectral regularization, we also implement a na\"ive biased estimator of the Hankel matrix for comparison defined by $\hat{\mH}=\sum\limits_{i=0}^{10}\mH^{(i)}_f$, which is  a fixed-sized subblock of the Hankel matrix. 
%, in which we merely account for its subblock in $\R^{\tilde{\Sigma}\times\tilde{\Sigma}}$ where $\tilde{\Sigma}:= \bigcup\limits_{i=0}^{10}\Sigma^i$.

\begin{figure}[H]
    \centering
   \includegraphics[width=1\textwidth]{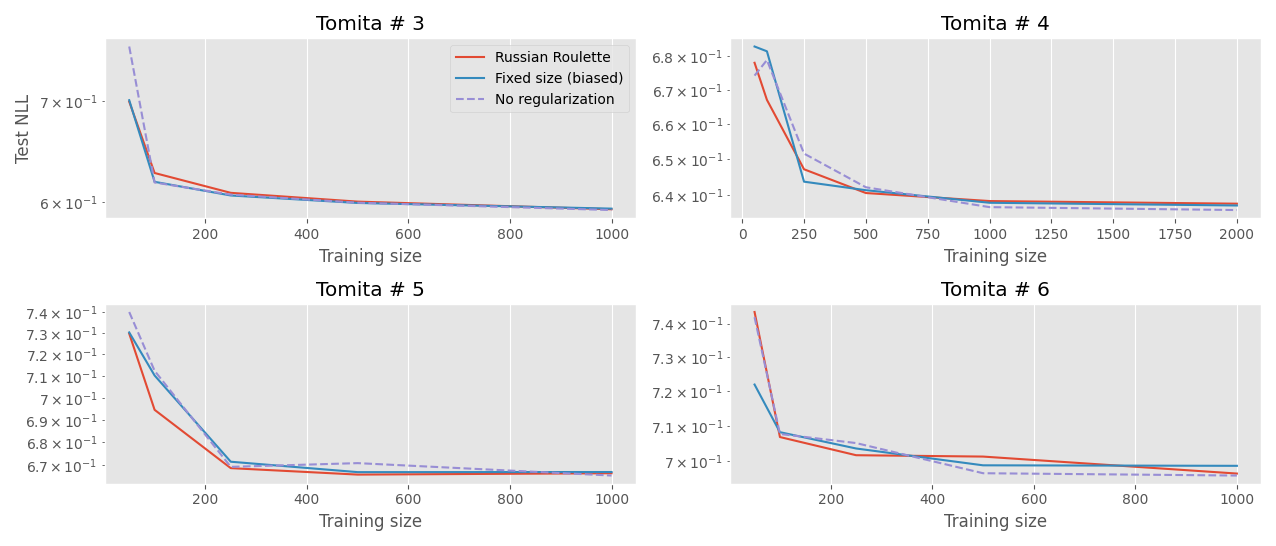}
    \caption{Test NLL against different training sizes for training with Russian Roulette estimator, with the fixed-sized (na\"ive biased) estimator, and without spectral regularization on Tomita grammars \#3 to \#6.}
    \label{fig:xp_tomita}
\end{figure}
Results on Tomita grammars \#3 to \#6 are presented in Figure~\ref{fig:xp_tomita}, where we see that spectral regularization marginally improves generalization for Tomita grammars \#4, \#5, and \#6 on small training data sizes. Especially on Tomita grammar \#5, the unbiased Russian Roulette estimator performs modestly better than the na\"ive biased one. However, there is no clear winner between the two estimators, biased or not, on other Tomita grammars. We hypothesize that this phenomenon might be due to the bias-variance tradeoff, i.e., the high variance of the unbiased Russian Roulette estimator makes the loss computation much coarser~\citep{DBLP:conf/icml/BeatsonA19}, which will be further investigated in future work. We also consider whether  more convincing results can be obtained on other tasks such as classification, or on other datasets, to be explored in upcoming studies.

\section{Conclusion}
%\gr{TODO!}
This paper proposes spectral regularization according to an intuitive notion of simplicity arising from the Chomsky hierarchy, which serves as an extra inductive bias for any sequence modeling task and is formulated as an additional regularization term to be added to any loss function. Results on synthetic data of Tomita grammars show that spectral regularization indeed marginally helps encourage the model to learn approximately low-rank functions. Forthcoming research will also examine the effect of spectral regularization in other tasks and other datasets.

To estimate the trace norm of the bi-infinite Hankel matrix in the spectral regularizer, we construct an unbiased stochastic estimator to relax the loss minimization problem. However, the unbiased estimator does not exhibit significant advantages compared to a na\"ive biased one, which will be explored in further research.

% Acknowledgements should go at the end, before appendices and references

\acks{We would like to acknowledge the support of the 2021 Globalink Research Internship Mitacs program (Project ID 24986).
% %\gr{TODO: add the MITAC project number if it exists} 
}


%\newpage

% \appendix
% \section*{Appendix A.}
% \label{app:theorem}

% Note: in this sample, the section number is hard-coded in. Following
% proper LaTeX conventions, it should properly be coded as a reference:

%In this appendix we prove the following theorem from
%Section~\ref{sec:textree-generalization}:

% In this appendix we prove the following theorem from
% Section~6.2:

% \noindent
% {\bf Theorem} {\it Let $u,v,w$ be discrete variables such that $v, w$ do
% not co-occur with $u$ (i.e., $u\neq0\;\Rightarrow \;v=w=0$ in a given
% dataset $\dataset$). Let $N_{v0},N_{w0}$ be the number of data points for
% which $v=0, w=0$ respectively, and let $I_{uv},I_{uw}$ be the
% respective empirical mutual information values based on the sample
% $\dataset$. Then
% \[
% 	N_{v0} \;>\; N_{w0}\;\;\Rightarrow\;\;I_{uv} \;\leq\;I_{uw}
% \]
% with equality only if $u$ is identically 0.} \hfill\BlackBox

% \noindent
% {\bf Proof}. We use the notation:
% \[
% P_v(i) \;=\;\frac{N_v^i}{N},\;\;\;i \neq 0;\;\;\;
% P_{v0}\;\equiv\;P_v(0)\; = \;1 - \sum_{i\neq 0}P_v(i).
% \]
% These values represent the (empirical) probabilities of $v$
% taking value $i\neq 0$ and 0 respectively.  Entropies will be denoted
% by $H$. We aim to show that $\fracpartial{I_{uv}}{P_{v0}} < 0$....\\

% \vskip 0.2in
\bibliography{literatures}

\end{document}